\documentclass[conference]{IEEEtran}
\IEEEoverridecommandlockouts

\usepackage[T1]{fontenc}
\usepackage[utf8]{inputenc}
\usepackage{graphicx}
\usepackage{amsmath}
\usepackage{amssymb}
\usepackage{booktabs}
\usepackage{cite}
\usepackage{amsmath,amssymb,amsfonts}
\usepackage{algorithm}
\usepackage{algorithmic}
\usepackage{graphicx}
\usepackage{textcomp}
\usepackage{xcolor}
\usepackage{comment}
\usepackage{booktabs}
\usepackage{multirow}
\usepackage{tcolorbox}
\usepackage{colortbl}
\usepackage{xcolor}

\definecolor{Rank1}{RGB}{200,255,200}  
\definecolor{Rank2}{RGB}{230,255,230}  
\definecolor{Rank3}{RGB}{255,255,200}  
\definecolor{Rank4}{RGB}{255,200,200}  
\usepackage[pagebackref,breaklinks,colorlinks]{hyperref}

\usepackage[capitalize]{cleveref}
\crefname{section}{Sec.}{Secs.}
\Crefname{section}{Section}{Sections}
\Crefname{table}{Table}{Tables}
\crefname{table}{Tab.}{Tabs.}

\begin{document}

\title{From Concept to Capability:\\ Evaluating 3D Gaussian Splatting for\\ Synthetic Scene Editing in Autonomous Driving}

\author{
\IEEEauthorblockN{
Ali Nouri\IEEEauthorrefmark{1}\IEEEauthorrefmark{4},
Yifei Zhang\IEEEauthorrefmark{2}\IEEEauthorrefmark{3},
Yifan Zhang\IEEEauthorrefmark{3},
Tayssir Bouraffa\IEEEauthorrefmark{1},\\
Zhennan Fei\IEEEauthorrefmark{1}\IEEEauthorrefmark{4},
Zijian Han\IEEEauthorrefmark{2},
Håkan Sivencrona\IEEEauthorrefmark{1},
Anders Heyden\IEEEauthorrefmark{3}
}

\IEEEauthorblockA{\IEEEauthorrefmark{1}Volvo Cars, Gothenburg, Sweden}

\IEEEauthorblockA{\IEEEauthorrefmark{2}Zenseact AB, Gothenburg, Sweden}

\IEEEauthorblockA{\IEEEauthorrefmark{3}Lund University, Lund, Sweden}

\IEEEauthorblockA{\IEEEauthorrefmark{4}Chalmers University of Technology, Gothenburg, Sweden\\
Email: ali.nouri@volvocars.com}
}

\begin{figure*}[t]
\centering
\begin{tcolorbox}[colback=yellow!10!white, colframe=yellow!50!black]
\centering \textbf{This is a preprint version. The final version will appear in the proceedings of SafeComp 2026.}
\end{tcolorbox}
\end{figure*}
\maketitle

\begin{abstract}
The perception of an Autonomous Driving System (ADS) critically depends on relevant, comprehensive, and diverse datasets to ensure its safety while operating in the environment. Field data collection lacks completeness with respect to the list of rare but still possible safety-related scenarios needed for the development, verification, and validation of the ADS.
3D Gaussian Splatting (3DGS) has shown promising capabilities for the reconstruction and editing of scenes based on data collected by cameras and LiDAR sensors. However, the industrial fidelity evaluation of reconstructions is underexplored, which is crucial when employing such methods in safety-related systems, especially for ADS. This becomes more challenging as ADS operates in a dynamic, uncontrolled environment with limited viewpoints and often partially occluded objects. This paper addresses this gap by proposing and implementing a framework (Fig.~\ref{fig:Overview}) to systematically analyze the capabilities and limitations of 3DGS for use in the reconstruction of safety-related scenes. It focuses on the quality of reconstruction for vehicles and pedestrians, which are the two most critical object classes for ADS.
Our findings provide industry insights into the fidelity degradation of reconstructions from multiple novel viewpoints, both lateral and longitudinal, enabling the integration of these methods into real-world industrial AD software development and testing pipelines.
\end{abstract}

\section{Introduction}
\label{sec:intro}

\begin{figure*}
  \includegraphics[width=\textwidth]{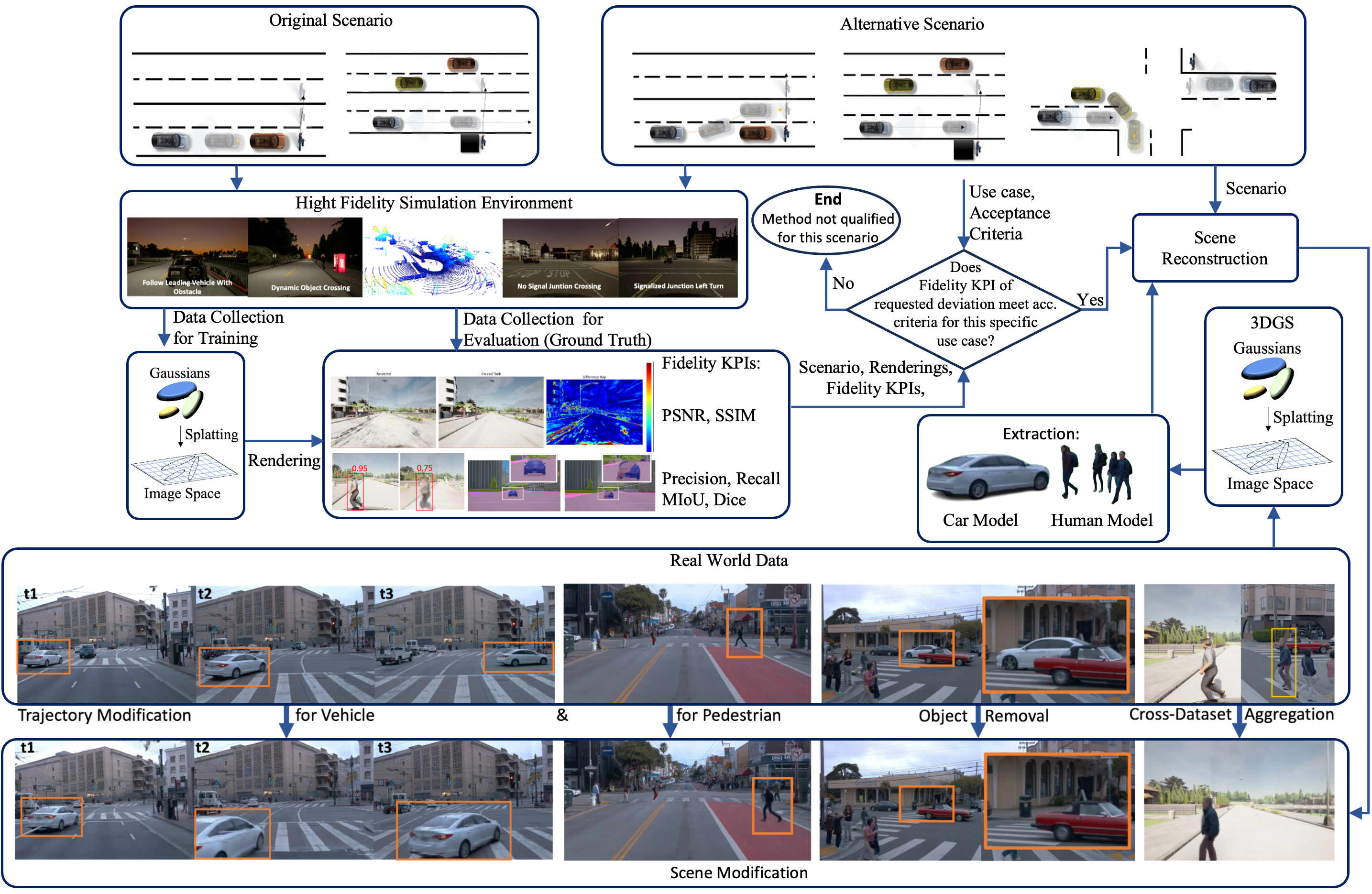}
  \caption{Presents the Dynamic Fidelity Evaluation framework for scene reconstruction: initially, data from the scene, both the original (for training) and alternative scenario (for evaluation), are collected from a simulation environment. Reconstruction is then performed based on the original view, and a novel viewpoint is rendered. The rendered data is compared against the evaluation data, fidelity KPIs are extracted, and finally, if the KPIs meet the acceptance criteria for each use case (e.g., AEB testing ``Avoid Collision'' during ''Cutting-in'' ), the reconstruction is considered valid. The bottom part presents multiple examples showcasing the capabilities of 3DGS in constructing alternative scenarios using real-world datasets. The top row shows the original data, while the bottom row displays the modifications. The two leftmost examples demonstrate 3DGS’s modification capabilities, while the two rightmost cases illustrate object removal (e.g., a white vehicle) and object injection (e.g., a pedestrian from real-world data into a simulation environment).}
  \label{fig:Overview}
\end{figure*}

ADS research and development traces back to the 1980s~\cite{gudla2022review}, but recent incidents, such as a mishap involving a robotaxi that resulted in severe pedestrian injury, highlight the persistent safety challenges in the field. In this case, the investigation revealed that while the ADS could detect the pedestrian and other agents separately, it failed to respond appropriately~\cite{cruise2024, nouri2025devsafeops}. This incident emphasizes the importance of the verification and validation (V\&V) of the whole system against rare and complex yet possible scenarios. Achieving higher safety requires innovative, high-fidelity simulation methodologies capable of scenario manipulation for effective testing. For instance, some scenarios are rarely encountered in the real world, such as animals on the road, while others could be too dangerous to collect, such as emergency collision data involving children or high-speed impacts. With a realistic simulation, all of these cases can be safely and massively generated within a reasonably short time frame. Moreover, a wide range of scenarios can be applied to ensure sufficient data diversity (the bottom of Fig.~\ref{fig:Overview}). These variations extend beyond environmental conditions like lighting, weather, or background, and also include differences in vehicle trajectories and viewpoints.
Moreover, the complexity of end-to-end transformer-based self-driving algorithms brings high requirements for data quality and volume~\cite{liu2024survey}. This is increasingly difficult to fully satisfy with only real-world vehicle-collected data, since the data collection process using test vehicles or fleets is slow and costly. However, the benefits of synthetic data in terms of efficiency and cost have shown great potential to address these issues. Data diversity is another crucial aspect to consider, since covering as many scenarios as possible during training is key to assure safety and robustness.
Using a game engine like Unity or Unreal Engine to generate synthetic data for training or testing can be seen as one potential solution. However, the purely game-engine-based method requires a huge amount of human effort for asset creation and environment tuning. It is also very challenging to ensure that manually created assets and environment settings are realistic enough for neural networks. As the need for different environments and fidelity levels increases, this method is very limited in terms of automation and scalability. Other data enrichment methods, such as GANs, can efficiently generate a large variety of new data but lack precise control over the data content.

Recent work in neural rendering methods, such as Neural Radiance Fields (NeRF) and 3D Gaussian Splatting (3DGS), underscores promising potential to address the aforementioned demand. 3DGS, in particular, has gained attention due to
its efficient reconstruction from camera and LiDAR data as well as real-time rendering performance, which is crucial for the training and V\&V of ADS software stacks. Despite these advancements, most studies evaluate 3DGS in specific scenarios without shedding light on its capabilities and limitations that must be considered when using it in a real industrial project. This is crucial for safety-related use cases, as the fidelity of the method’s output must be considered, and arguments around this aspect are often the missing part of the process.
Hence, this paper aims to critically assess the capabilities and limitations of 3DGS and proposes a framework to systematically define the boundaries based on acceptable fidelity metrics. In this work, we mainly focus on scenarios involving vehicles (high exposure) and pedestrians (high exposure and high severity). Additionally, we integrate multiple methods to improve the reconstruction of pedestrians within the framework. By rigorously identifying the fidelity boundaries of 3DGS, this research aims to serve as an industrial guideline for stakeholders, highlighting the capabilities and limitations of synthetic data generation and providing clear guidance for the effective integration of 3DGS into ADS development and V\&V.

\section{Related Work}
\label{sec:relatedwork}

\textbf{Data gap and simulation:} Despite the significant increase in the quality and variety of autonomous driving datasets in recent years, real-world data often proves insufficient and not inclusive of all unknown or complex scenarios~\cite{liu2024survey}.
Moreover, some scenarios are the result of analysis methods, such as Hazard Analysis and Risk Assessment~\cite{Nouri2024CAINLLM}, which lead to requirements and then to implemented algorithms. However, when moving toward end-to-end approaches, it becomes the role of data to represent these scenarios, which are difficult to capture because they are hypothetical.

Synthetic data generation by employing computer graphics and simulations is proposed as a potential cost-effective and flexible approach to improve diversity in datasets and construct difficult or unsafe scenarios~\cite{song2023synthetic}. Several simulation environments have been developed for ADS, such as  LGSVL~\cite{rong2020lgsvl} and Car Learning to Act (CARLA)~\cite{dosovitskiy2017carla}. For instance, CARLA is an Unreal Engine-based simulation environment focused on urban environments with customizable weather conditions. However, data-driven methods are gaining attention, as assets in simulation environments are not easily scalable or realistic enough~\cite{chen2024omnire}, especially when it comes to tackling previously rare or unseen objects.

\textbf{NeRF vs. 3DGS:} Prior work on neural scene reconstruction for view synthesis in driving environments largely follows two families: volumetric, NeRF-style methods and explicit, 3D Gaussian Splatting (3DGS). NeRF represents a scene as an implicit radiance field and achieves high-quality novel views from sparse, real-world imagery~\cite{mildenhall2021nerf}. However, the combination of per-scene optimization and volumetric ray marching leads to comparatively high compute cost for both training and inference, which becomes a practical limitation for autonomous driving where large volumes of re-constructed data are needed for development, verification,
and validation~\cite{hess2025splatad}. Even with accelerated variants, end-to-end reconstruction typically remains minutes to days per scene depending on hardware, resolution, and view coverage, and city-scale deployments commonly partition space into many sub-models, increasing operational overhead.
3DGS addresses these efficiency constraints by replacing volumetric integration with explicit anisotropic 3D Gaussians rendered via differentiable splatting/rasterization, while learning per-primitive appearance and opacity~\cite{kerbl20233d}. Under common settings, this formulation yields interactive to real-time rendering and shorter time-to-usable reconstructions, with competitive perceptual quality relative to NeRF-style renderers~\cite{hess2025splatad, kerbl20233d}. 
For AD/ADAS use cases, 3DGS’s runtime profile is better aligned with industrial needs, since perception-in-the-loop evaluation demands high frame throughput, frequent re-rendering from novel viewpoints, and coverage of large outdoor environments. Beyond camera-only setups, recent 3DGS pipelines also integrate multi-sensor inputs (e.g., LiDAR) to stabilize geometry and reduce view-dependent artifacts, which is valuable for safety-critical analysis and editing workflows~\cite{hess2025splatad, kerbl20233d}.

Within the 3DGS family tailored to road scenes, StreetGS~\cite{yan2024street} is a specialized implementation of the 3D Gaussian Splatting approach to reconstruct both static and dynamic elements of street scenes from multiple camera views. To improve the fidelity of specific road users, such as pedestrians and cyclists, and to capture details of human posture and movement, methods including OmniRe~\cite{chen2024omnire}, SplatAD~\cite{hess2025splatad}, and DistillNeRF~\cite{wang2024distillnerf} have been introduced. Taken together, 3DGS currently offers a more practical balance of fidelity, efficiency, and ecosystem compatibility for industrial-scale AD/ADAS pipelines, which motivates our focus on 3DGS-based benchmarking in this work.

\textbf{Pedestrian gesture modelling:} As pedestrians are involved in scenarios with the highest risk (i.e., high severity in the event of an incident and high exposure in driving scenarios), specialized human modeling and generation methods are required to improve the fidelity of the reconstructed scenes. Human body gestures are an important input for perception, as they affect pedestrian trajectory prediction, which can be used in the precautionary behavior of both AD and ADAS functions. The complex motion characteristics and appearances of the human body require specialized modeling techniques, which impose limitations on the aforementioned methods.
Skinned Multi-Person Linear Model (SMPL)~\cite{loper2023smpl} improves human reconstruction by incorporating the known human body structure, which is further enhanced by NeuMan~\cite{jiang2022neuman}.

\textbf{Research gap:} However, none of the above studies systematically evaluated the performance of 3DGS for different novel viewpoints against the ground truth to demonstrate the capabilities of the models in a dynamic environment, such as the one in our application, which has limited viewpoints.There is a lack of comprehensive industrial guidelines for employing 3DGS in safety-related software development in the automotive domain, and more specifically in AD development and testing, which will be addressed in this study. This includes acceptable deviations from the original viewpoint and the use of different fidelity metrics for the output. In this study first we captured the capabilities and limitations of 3DGS-based method, then specified requirements on inputs of each method, and finally combined them to get the most out of them.

\section{Methodology}
\label{sec:methodology}

This section describes the methodological framework designed to identify the capabilities and limitations of 3DGS by evaluating its fidelity and applicability under different spatial deviations from the original viewpoint, as illustrated in Figure~\ref{fig:Overview}. Our approach includes systematic data acquisition for both the original viewpoint and the ground truth of novel viewpoints to accurately evaluate reconstruction performance. We first compared state-of-the-art methods, then focused on the best-performing model, assessing its capabilities across different object classes.
Finally, we integrated complementary methods to address its identified limitations.

\subsection{3D Gaussian Splatting Reconstruction}

We implemented a pipeline employing 3DGS, and among various candidate models, we selected DeformGS, StreetGS, PVG, and OmniRe. We then focused further on OmniRe, as it outperforms the others.
Our pipeline uses OmniRe, a dynamic 3DGS method, to reconstruct scenes from multi-modal simulation data. OmniRe organizes the scene as a hierarchical Gaussian Scene Graph, modeling static backgrounds, vehicles, and pedestrians with distinct sets of Gaussians. Inputs include calibrated RGB images and LiDAR point clouds from CARLA, with preprocessing to extract camera parameters and segment dynamic objects. During training, OmniRe optimizes the spatial, visual, and transparency attributes of the Gaussians to minimize rendering error. By integrating LiDAR depth, it enhances convergence speed and spatial precision, particularly around object boundaries. We also introduce training augmentations like camera perturbations and spatial regularization to improve generalization and prevent overfitting to training trajectories.

Due to the complexity and non-rigid behavior of humans, these methods lack adequate performance in modeling pedestrians. Given their criticality, we subsequently employed more specialized models for human body reconstruction~\cite{loper2023smpl}.
Our framework leverages the Gaussian scene representation to enable real‑time, training‑free edits of reconstructed environments. By directly manipulating the Gaussian nodes, we can remove or translate rigid objects, simply omitting a vehicle’s Gaussians to delete it, or applying a rigid-body transform like translation and rotation to reposition it along a new trajectory. 
These operations let us synthesize safety‑related scenarios, like a jaywalking pedestrian suddenly dashing into the road or an oncoming car veering into the ego‑lane, while preserving photorealism and physical consistency. Finally, a lightweight trajectory module, such as straight interpolation, S‑curve, lane‑change, produces novel-view camera paths across the edited scenes, yielding the tailored synthetic sequences for downstream evaluation.

\subsection{Data Acquisition from Simulation}
CARLA (Car Learning to Act), a high-fidelity simulation environment, is used to generate a diverse dataset by altering only one factor per scenario. CARLA is an open-source simulation platform built on Unreal Engine 4, capable of recreating various weather conditions (e.g., clear day, rain, sunset). It supports multiple sensor modalities, such as cameras and LiDAR, and automatically provides detections and ground truth data, including semantic segmentation.
CARLA offers several key advantages that make it an ideal for evaluating reconstruction methods:

\textbf{Controlled Environment:} CARLA enables precise control over agents in the scene, lighting conditions, and environmental factors, which is essential for eliminating confounding variables.

\textbf{Novel Viewpoint Sampling:} Unlike real-world datasets, which are restricted to sensors mounted on the vehicle, CARLA allows for reconstruction of scenarios from novel viewpoint trajectories, which can be used as ground truth for evaluating pipeline output. Moreover, it enables data collection from sensors placed outside the vehicle.

\textbf{Accurate and Effortless Ground Truth:} The simulator provides instant, accurate ground truth for all modalities (RGB, depth, semantics) from any viewpoint, facilitating comprehensive evaluation. This allows us to test various scenarios without the burden of manually generating ground truth and eliminates inaccuracies in the evaluation process.

\subsection{Real World Datasets}
In this study we have selected Waymo Open Dataset~\cite{sun2020scalability}, Zenseact Open Dataset~\cite{alibeigi2023zenseact}, and KITTI~\cite{geiger2013vision} due to their high-quality LiDAR and vision sensor data with fused data and detailed annotations, collected across various driving conditions. We showcased the advantage of the method by extracting a human from one dataset and injecting it into another scenario in a different dataset to demonstrate the method’s capability for aggregating object models, an essential feature for industrial use cases, since datasets are sometimes collected by different parties with different sensor setups. Moreover, it enables the industry to share their datasets and treat the objects in them like assets in simulation, which helps with sharing rare agents or behaviors across the industry to better collaborate and improve system performance.

\subsection{Experimental Setup}
We conduct experiments on a suite of four simulated driving scenarios created in
CARLA. These scenarios include: (1) Following: the ego vehicle following a lead car on a
straight road with a pedestrian on the sidewalk, (2) Intersection: the ego car approaching an intersection
where cross-traffic and pedestrians are present, (3) Overtaking: a scenario with a stalled vehicle and the
ego vehicle changing lanes to overtake, and (4) Oncoming: a two-way road scenario with an oncoming car
that potentially swerves. Each scenario is run for a few seconds, yielding on the order of 200 multi-view
frames (1000 images across the 5 cameras). The multi-view images and LiDAR point clouds from each
scenario are used to train four separate 3DGS reconstructions. We evaluate several 3D reconstruction
methods on these scenarios for comparison: DeformGS \cite{duisterhof2023deformgs} as a baseline image-based
dynamic NeRF, PVG \cite{chen2023periodic} which uses explicit 3D bounding boxes to guide Gaussian placement,
StreetGS \cite{yan2024street} which incorporates LiDAR for street scenes, and OmniRe\cite{chen2024omnire} - which
uses LiDAR and a scene graph. We use the authors' released implementations for these methods and apply
a consistent set of parameters and training iterations for fairness. All reconstructions are supervised with
the same set of input views and LiDAR, when applicable. After training, for each method and scenario, we
render novel views not seen during training. We define a systematic set of camera poses for novel view
testing, which includes view-points along the original trajectory to test interpolation and increasingly distant
viewpoints that deviate laterally and vertically from the path to test extrapolation. We categorize novel
views into three bins: near (0.5m from training path), medium (1.6m), and far (3.2m). \footnote{Our implemented framework is available at \href{https://github.com/anonWACV1/3D-Reconstruction/tree/main}{GitHub repository}}  

\subsection{Metrics used for evaluation}
\label{sec:methodologyMetrics}
To rigorously assess the fidelity of scene reconstruction by each method, we designed a layered evaluation strategy that combines state-of-the-art computer vision metrics with AD perception-oriented measures. This breakdown enables us to better analyze the capabilities and limitations of the methods in reconstructing different objects and aspects of a scene for both original and novel viewpoints. 
Peak Signal-to-Noise Ratio (PSNR) \cite{chen2024omnire} and Structural Similarity Index Measure (SSIM) \cite{chen2024omnire} are used in computer vision to quantitatively evaluate rendering quality compared to the ground truth. In our study, these two metrics are computed for both the full image and specifically for non-sky regions and dynamic objects (vehicles and pedestrians).
However, PSNR and SSIM alone may not fully capture the reconstruction fidelity from the perspective of AD perception. To address this, two state-of-the-art robust object detection models, YOLOv10~\cite{wang2024yolov10} and DETR~\cite{carion2020end}, are used to assess how detectable the modified agents are in the scene. Both models are real-time CNN-based detectors that are pre-trained on the COCO dataset, which avoids data leakage from our datasets (CARLA and real datasets). For each rendered novel view, we evaluate detection precision and recall for full images, vehicles, and pedestrians.
Lastly, semantic segmentation metrics are calculated to evaluate the quality of pixel-level semantic understanding, which is a crucial capability for safe path planning and collision avoidance. To this end, mean Intersection-over-Union (mIoU)\cite{xie2021segformersimpleefficientdesign} (i.e., an indicator of average overlap) and Dice coefficient \cite{jadon2020survey} (i.e., an indicator of object boundaries) scores are used, which complement the object detection evaluations.

\begin{table*}
\centering
\caption{Comparison of different methods on scene reconstruction and average of spatial novel view for 10 cameras over three different lateral distances. The heatmap highlights relative rankings, from green to red indicating best to worst. StreetGS and DeformGS failed to reconstruct human (as they cannot model deformable objects~\cite{chen2024omnire}); hence, the corresponding metrics are marked as not applicable (N/A).}
\resizebox{\textwidth}{!}{
\begin{tabular}{lccccccccccccccc}
\toprule
\multirow{4}{*}{Methods} & \multirow{4}{*}{Box} & \multirow{4}{*}{LiDAR} & \multicolumn{6}{c}{Scene Reconstruction} & \multicolumn{6}{c}{Spatial Novel View} \\
\cmidrule(lr){4-9} \cmidrule(lr){10-15}
& & & \multicolumn{2}{c}{Full Image} & \multicolumn{2}{c}{Human} & \multicolumn{2}{c}{Vehicle} & \multicolumn{2}{c}{Full Image} & \multicolumn{2}{c}{Human} & \multicolumn{2}{c}{Vehicle} \\
\cmidrule(lr){4-5} \cmidrule(lr){6-7} \cmidrule(lr){8-9} \cmidrule(lr){10-11} \cmidrule(lr){12-13} \cmidrule(lr){14-15}
& & & PSNR$\uparrow$ & SSIM$\uparrow$ & PSNR$\uparrow$ & SSIM$\uparrow$ & PSNR$\uparrow$ & SSIM$\uparrow$ & PSNR$\uparrow$ & SSIM$\uparrow$ & PSNR$\uparrow$ & SSIM$\uparrow$ & PSNR$\uparrow$ & SSIM$\uparrow$ \\
\midrule
DeformGS(Yang et al., 2023c) & $\checkmark$ & & \cellcolor{Rank4}27.82 & \cellcolor{Rank4}0.915 & N/A & N/A & \cellcolor{Rank3}25.18 & \cellcolor{Rank2}0.860 & \cellcolor{Rank2}22.08 & \cellcolor{Rank2}0.733 & N/A & N/A & \cellcolor{Rank3}20.92 & \cellcolor{Rank4}0.732 \\
PVG(Chen et al., 2023) & $\checkmark$ & & \cellcolor{Rank2}33.32 & \cellcolor{Rank2}0.942 & \cellcolor{Rank2}19.41 & \cellcolor{Rank2}0.621 & \cellcolor{Rank4}24.23 & \cellcolor{Rank3}0.812 & \cellcolor{Rank4}20.04 & \cellcolor{Rank4}0.649 & \cellcolor{Rank1}15.33 & \cellcolor{Rank1}0.645 & \cellcolor{Rank4}20.24 & \cellcolor{Rank3}0.743 \\
StreetGS(Yan et al., 2024) & $\checkmark$ & $\checkmark$ & \cellcolor{Rank3}31.84 & \cellcolor{Rank3}0.937 & N/A & N/A & \cellcolor{Rank2}26.18 & \cellcolor{Rank4}0.808 & \cellcolor{Rank3}20.66 & \cellcolor{Rank3}0.656 & N/A & N/A & \cellcolor{Rank2}21.43 & \cellcolor{Rank2}0.789 \\
OmniRe(Chen et al., 2024) & $\checkmark$ & $\checkmark$ & \cellcolor{Rank1}34.07 & \cellcolor{Rank1}0.951 & \cellcolor{Rank1}23.90 & \cellcolor{Rank1}0.787 & \cellcolor{Rank1}28.30 & \cellcolor{Rank1}0.876 & \cellcolor{Rank1}23.75 & \cellcolor{Rank1}0.768 & \cellcolor{Rank2}15.16 & \cellcolor{Rank2}0.619 & \cellcolor{Rank1}21.52 & \cellcolor{Rank1}0.806 \\
\bottomrule
\end{tabular}
}
\label{tab:comparisonPSNRSSIMAllMethod}
\end{table*}

\subsection{Pass/Fail criteria for reconstruction fidelity:}

Reconstruction techniques such as 3DGS are relatively new and fidelity guidelines are still missing in standards such as ISO 26262~\cite{ISO26262}, 21448~\cite{sotif}, and 4804~\cite{isotr4804}. However, the required guidelines for these new approaches can be extrapolated by looking at guidelines for methods with similar purposes. For instance, according to ISO 4804~\cite{isotr4804}, the model used in a simulation environment can be accurate only to a certain degree, and the pass/fail criteria for accuracy highly depend on the test goal. For example, when testing the behavior of a function, a lightweight and abstract simulation environment could be sufficient, whereas for other tests, even a high-fidelity simulation environment may not suffice, and real-world testing is required.
By adapting the process described in Part 8 of ISO 26262~\cite{ISO26262}, which addresses conventional tools and methods used in the development of safety-related software, the pass/fail criteria can be defined based on the following factors:

\textbf{Reconstruction error impact:} As reconstructed data can be used for both training and testing of AD software, it is crucial to distinguish between these use cases. Using reconstructed data for training could contaminate the software deployed in the vehicle, making the impact significantly higher than when the same data is used for verification and validation.

\textbf{Reconstruction error detection:} Errors in synthetic data can often be detected, especially when the software is tested in simulation environments using various available methods. However, detecting the effects of errors in training data is more difficult, and even when such effects are identified, it is hard to trace them back to the training data.

\textbf{Safety Requirements:} Safety requirements of the perception system are further broken down into data quality requirements. For instance, in Automated Emergency Braking (AEB), false positives (i.e., brake requests when there is no imminent collision) are assigned a higher Automotive Safety Integrity Level (ASIL) than false negatives (i.e., no brake request when a collision is imminent). This is because, in the case of a false negative, the driver can still react and apply the brakes, resulting in a lower ASIL. However, if the system brakes unnecessarily, the driver may not have enough time to respond, potentially causing a rear-end collision. Therefore, in this context, precision should be prioritized, while lower recall is more acceptable. The criticality of these requirement changes in AD, where both false positives and false negatives must be minimized.

Hence, as shown in Figure~\ref{fig:Overview}, the fidelity of reconstructions should be evaluated for each specific scenario and use case. In this study, we assume that if the drops in precision or recall are less than 10\%, the synthetic images can be considered to have acceptable fidelity.

\section{Results \& Discussion}
\label{sec:results_discussion}
\subsection{Quality of The Spatial Novel View}

\begin{figure}
  \includegraphics[width=\columnwidth]{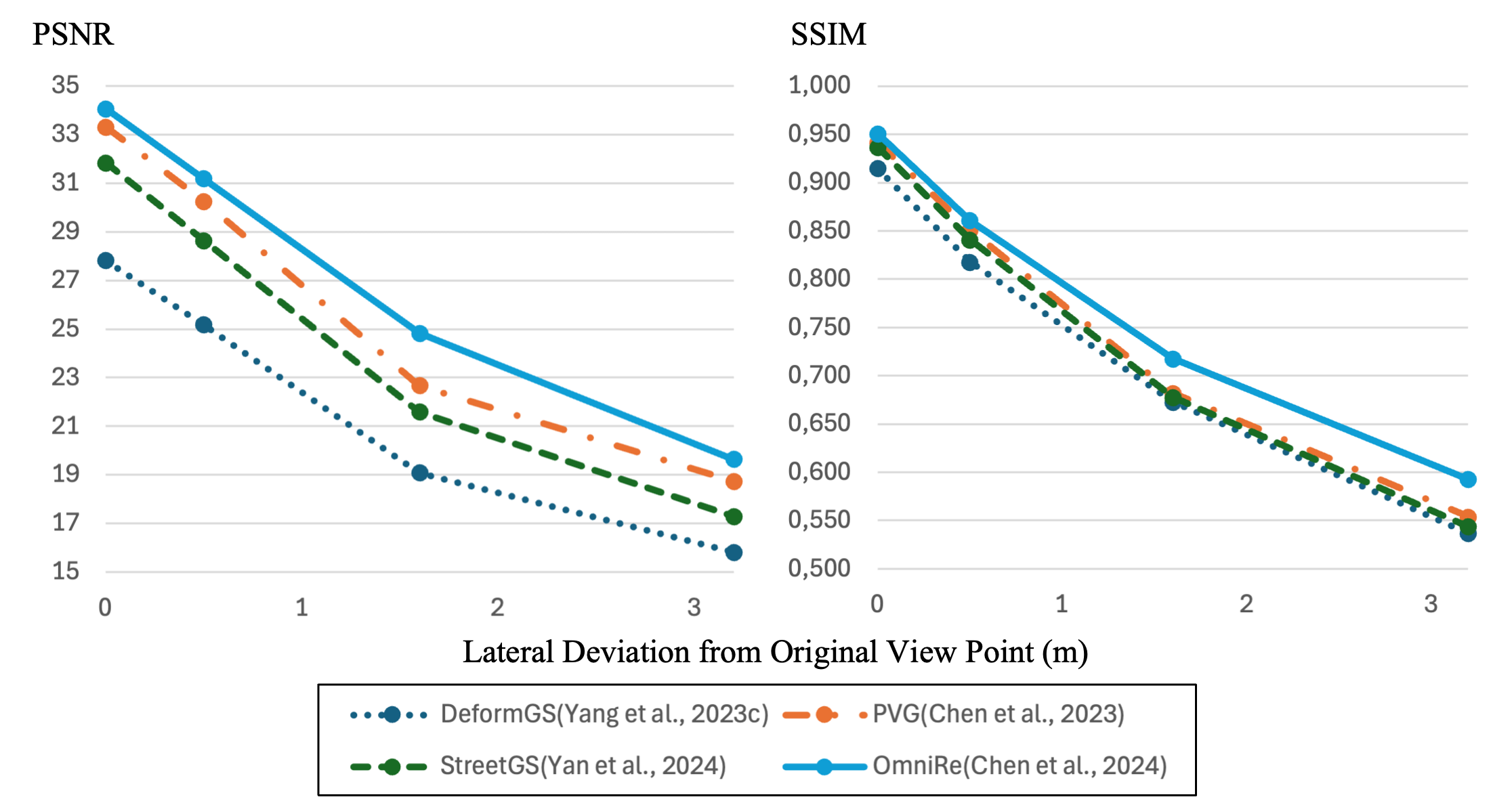}
  \caption{Evaluation of four methods based on distance for PSNR and SSIM metrics. As it is presented the performance of all methods drop by increasing spatial distance in novel view point. Moreover, OmniRe consistantly has better performance compared with other three methods.}
  \label{fig:PSNRSSIMAllMethods}
\end{figure}

As shown in Table~\ref{tab:comparisonPSNRSSIMAllMethod}, OmniRe consistently demonstrates the best performance for both original and novel viewpoint reconstruction with respect to PSNR and SSIM for the full image. When looking at specific object classes, StreetGS and DeformGS show the weakest performance for human reconstruction in both original and novel viewpoints, while StreetGS achieves relatively strong performance for vehicles, ranking second after OmniRe. A potential reason for this might be StreetGS’s dependency on LiDAR information, which is normally insufficient for humans. The non-rigid nature of the human body requires a high point cloud density, which is generally not available, whereas vehicles, due to their rigid structure, require less.
This pattern confirms that different scene elements may require tailored reconstruction strategies. Moreover, OmniRe presents the strongest performance for both pedestrians and vehicles. The results also indicate that OmniRe and StreetGS, which use LiDAR data, achieve overall better performance compared to DeformGS and PVG, which do not use it.

The performance evaluation of the methods is further visualized in Figure~\ref{fig:PSNRSSIMAllMethods} for different lateral distances from the training trajectory, which shows that all methods experience a clear and consistent drop in PSNR and SSIM. OmniRe consistently maintains higher performance, while the other methods rank next in overall performance, with PVG, StreetGS, and DeformGS following in order.
This leads to the conclusion that spatial generalization is fundamentally limited in current 3DGS pipelines, particularly for scenes with dynamic objects and limited viewpoints on objects, which are critical for safe autonomous driving.
Furthermore, depending on the required fidelity limit, the reconstruction validity is bounded to a certain lateral distance from the training trajectory.

\begin{figure}
  \includegraphics[width=\columnwidth]{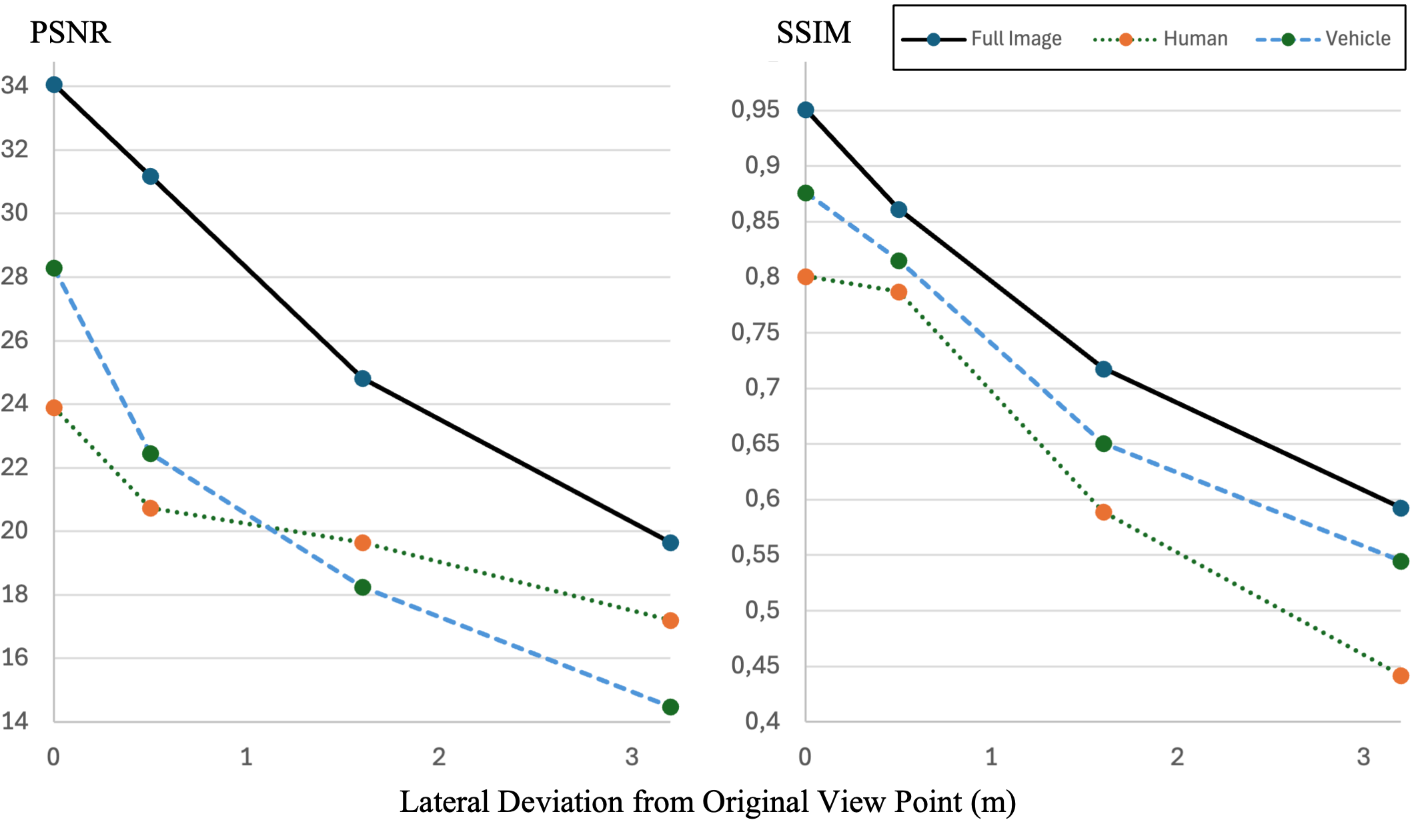}
  \caption{Evaluation of OmniRe across spatial distances and object categories. The performance drops as the novel viewpoint goes further, and it is the same trend for the two main dynamic objects (human and vehicle).}
  \label{fig:PSNRSSIMALLObjectsOmniRe}
\end{figure}

OmniRe, as the best performer, is selected for further investigation of how the performance for humans and vehicles differs with respect to the full image, as shown in Figure~\ref{fig:PSNRSSIMALLObjectsOmniRe}. For these two dynamic objects, the performance is understandably lower, while the drop in performance with increasing lateral distance follows the same slope as for the full image. This highlights that human and vehicle reconstruction requires denser data collection and, if possible, complementary generative techniques to maintain high-fidelity representations in novel views.
By looking more closely at Figure~\ref{fig:PSNRSSIMALLObjectsOmniRe}, it can be seen that PSNR is slightly higher for humans than for vehicles, while for SSIM, the opposite is true. This suggests that while pixel intensities are more accurate (higher PSNR) for humans compared to vehicles, the structural details are less accurate (lower SSIM). This is a result of the rigidity of vehicles versus the dynamic and softer structure of pedestrians. This highlights the importance of considering both metrics together, as they capture complementary aspects of reconstruction quality.

\subsection{Object detection metrics for rendered images}

\begin{table}
\centering
\caption{Comparison of different methods on average precision and recall by Yolo and DETR.}
\resizebox{\columnwidth}{!}{
\begin{tabular}{lcccccc}
\toprule
\multirow{2}{*}{Methods} & \multicolumn{2}{c}{Full Image} & \multicolumn{2}{c}{Human} & \multicolumn{2}{c}{Vehicle} \\
\cmidrule(lr){2-3} \cmidrule(lr){4-5} \cmidrule(lr){6-7}
& Precision$\uparrow$ & Recall$\uparrow$ & Precision$\uparrow$ & Recall$\uparrow$ & Precision$\uparrow$ & Recall$\uparrow$ \\
\midrule
DeformGS & \cellcolor{Rank4}0.893 & \cellcolor{Rank3}0.866 & N/A & N/A & \cellcolor{Rank4}0.931 & \cellcolor{Rank3}0.893  \\
PVG       & \cellcolor{Rank3}0.924 & \cellcolor{Rank4}0.654 & \cellcolor{Rank2}0.858 & \cellcolor{Rank2}0.566 & \cellcolor{Rank1}0.978 & \cellcolor{Rank4}0.726 \\
StreetGS   & \cellcolor{Rank2}0.931 & \cellcolor{Rank1}0.931 & N/A & N/A & \cellcolor{Rank2}0.970 & \cellcolor{Rank2}0.968 \\
OmniRe   & \cellcolor{Rank1}0.943 & \cellcolor{Rank2}0.915 & \cellcolor{Rank1}0.919 & \cellcolor{Rank1}0.833 & \cellcolor{Rank3}0.963 & \cellcolor{Rank1}0.982\\
\bottomrule
\end{tabular}
}
\label{tab:comparisonPrecisionReecallAllMethods}
\end{table}

\begin{figure*}
  \includegraphics[width=\textwidth]{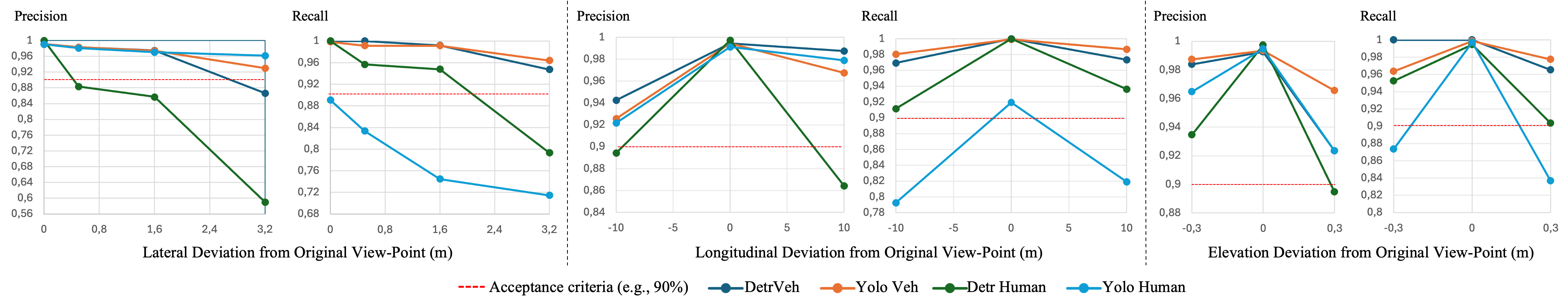}
  \caption{Presents the evaluation of OmniRe across multiple spatial distances (lateral, longitudinal, and elevation) and object categories, based on precision and recall for DETR and YOLO object detection. The ranges for each direction are selected based on the maximum required deviations from the original viewpoint: 3.2m for lateral (e.g., lane changes), 10m for longitudinal, and 30cm for elevation (e.g., due to road elevation or sensor placement on different platforms). As shown, the performance drop in the lateral direction is significantly greater than in the longitudinal direction, even for smaller deviations.}
  \label{fig:PrecisionRecalAllMethods}
\end{figure*}

According to Table~\ref{tab:comparisonPrecisionReecallAllMethods}, the precision and mAP across the four reconstruction methods, which helps to better measure the fidelity of each reconstructed object class from the perspective of an object detection algorithm (i.e., YOLOv10 and DETR). OmniRe on average outperforms the other methods in both human and vehicle object categories for both YOLO and DETR.
Comparing the detection metrics highlights the higher fidelity for vehicle classes due to the rigidity of vehicles and their well-defined geometry in reconstruction. On the contrary, the detection performance for humans is significantly lower for PVG. StreetGS and DeformGS fail to reconstruct the human at all, as they cannot model deformable objects~\cite{chen2024omnire}, which is a significant drawback for AD applications.
Moreover, while OmniRe’s performance for humans is better than that of the other methods, it is still lower than for vehicles, which is also reflected in the computer vision metrics.
Figure~\ref{fig:PrecisionRecalAllMethods} presents OmniRe's performance for vehicle and human reconstruction under deviations in different directions, highlighting that the performance drop in lateral deviations is much greater than in longitudinal ones.

\begin{figure}
  \includegraphics[width=\columnwidth]{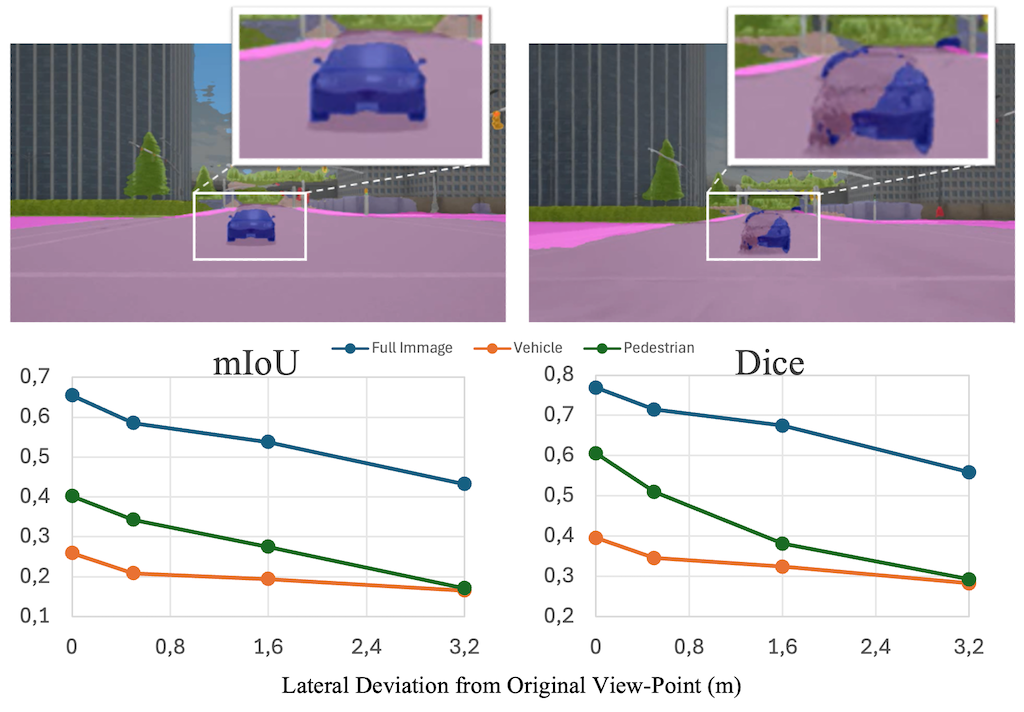}
  \caption{The top left figure shows novel view with 0.5 m spatial difference with relatively good segmentation quality where the vehicle is correctly identified and segmented with clear boundaries; the right side is presenting 3.2 m spatial difference in novel view showing significant degradation in segmentation performance. The bottom part is presenting the evaluation of SegFormer on OmniRe reconstruction across spatial distances and object categories.}
  \label{fig:SegFormer2}
\end{figure}

\subsection{Semantic Segmentation Performance}
As presented in Figure~\ref{fig:SegFormer2}, both mIoU and Dice coefficient degrade for OmniRe reconstructions with increasing spatial distance from the original training trajectory. mIoU and Dice remain relatively high for the full image, which includes both static and dynamic objects, while they show far steeper declines for vehicles and humans as the two dynamic objects. These sharp drops in Dice coefficients confirm the structural breakdown in pixel-level understanding.
These results confirm that while OmniRe maintained overall consistency with increasing distance, its ability to represent accurate semantics for dynamic objects drops significantly at greater distances.

\subsection{Viewpoint Coverage \& Reconstruction Validity}
\begin{figure}
  \includegraphics[width=\columnwidth]{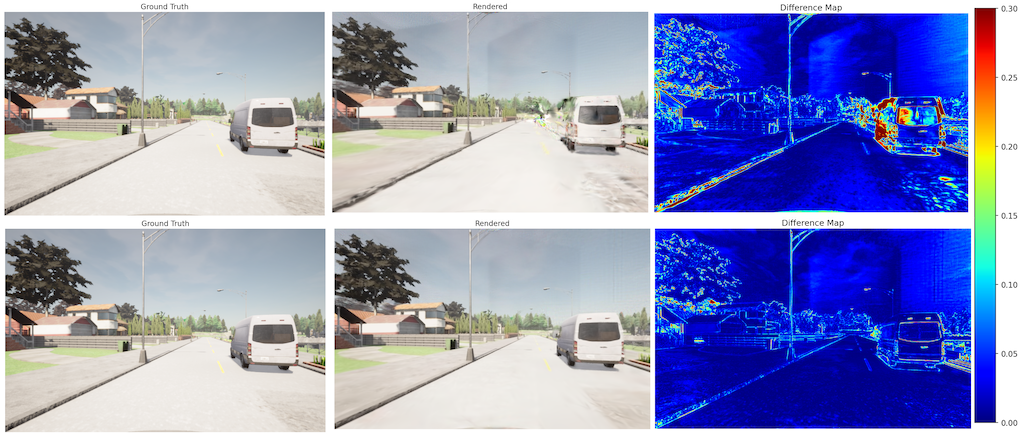}
  \caption{Presents a difference map for the same scenario with a vehicle that is trained using data from different angles. The top example is trained only with data from behind, while the bottom one is trained with data from behind and the side. This illustrates why each object should be tagged with valid reconstruction angles to ensure acceptable fidelity.}
  \label{fig:DifMap2ObjectsDifferentCollections}
\end{figure}

Difference maps presented in Figure~\ref{fig:DifMap2ObjectsDifferentCollections} show the reconstruction quality for the same object with different training data coverage from various viewpoints. The top example illustrates a vehicle reconstructed using only rear-view training data, while the bottom example shows the same vehicle reconstructed from a separate scenario with more comprehensive viewpoint coverage. This highlights the importance of assigning explicit visibility tags to each object, indicating which angles have sufficient data coverage and are therefore valid for novel view synthesis to satisfy fidelity requirements.
This is crucial because, in AD and ADAS, unlike other domains, it is not feasible to collect perfect 360° data for every object. Instead, reconstruction must be done using data with different levels of visibility while still maintaining fidelity. By automatically identifying objects with robust coverage, the pipeline can ensure that only well-supported sides of an object are used in reconstruction.

\subsection{Scene Editing and Edge Case Generation}

So far, we have established the capabilities and limitations of each reconstruction method by analyzing spatial generalization, especially for humans and vehicles, as the two most important object classes in our application. As the final step, the current section demonstrates how our scene editing pipeline can be applied to real-world datasets to ensure visual consistency. Crafting safety-related scenarios from data recorded in the field with an acceptable level of fidelity requires the combination of multiple agents and modifications to their trajectories. These agents include both rigid (e.g., vehicles) and non-rigid (e.g., pedestrians) objects.
The bottom of Fig.~\ref{fig:Overview} demonstrates scene editing applied to real-world datasets. The figure illustrates the lateral shifting and change in speed of a vehicle, effectively transforming one normal scenario into a safety-related cut-in scenario. In another example, the pedestrian’s trajectory is shifted from walking normally on a zebra crossing to a "jay-runner" scenario near the ego vehicle, which is more critical.

As rare scenarios and objects might not be captured by a single developer or organization, they can be extracted and shared with other parties to test their software across companies or be used in simulation environments as replacements for manually modeled assets. Hence, we extended our pipeline to demonstrate cross-dataset agent injection, which is showcased in the bottom of Fig.~\ref{fig:Overview}.

\subsection{Time and resource usage}

All experiments (3DGS training and rendering) were run on a workstation with a single NVIDIA RTX~6000 Ada Generation GPU (48~GB VRAM), Intel\textsuperscript{\textregistered} Core\textsuperscript{\texttrademark} i9-10920X CPU @ 3.50~GHz (24~cores), and 64~GB RAM, running CUDA~12.2 with NVIDIA driver~535.247.01.
Data were collected in \textit{CARLA} at 15~FPS with five cameras at $375 \times 1242$ resolution. We use four 8-second scenarios (each with 120~frames per camera, 600~frames total). Peak GPU memory during training was 40~GB of 48~GB for all scenarios. Per-scenario training times were 00:46:47, 00:48:17, 00:47:32, and 00:41:53 (hh:mm:ss), averaging 46~min~07~s per 8-second scenario at the above settings. For rendering (image to video at 15~FPS), on average, a representative scenario took 6~s with $<20$~GB GPU VRAM and 7.4~GB RAM.
The measurements above illustrate two practical advantages of 3D Gaussian Splatting for AD/ADAS pipelines. First, training cost is tractable on a single machine: sub-hour per scenario with one high-VRAM GPU (no cluster or multi-GPU orchestration), which makes scaling across many short scenarios a straightforward batch process and enables rapid re-training after edits. Second, inference and rendering are fast: 21~FPS at the stated resolution with modest GPU utilization and $<20$~GB VRAM, leaving significant headroom for concurrent processes (e.g., perception-in-the-loop evaluation, multi-view rendering, or on-the-fly scenario edits). Together, these properties translate to high throughput and low operational overhead, aligning with industrial needs to generate and re-render large volumes of camera trajectories for development, V\&V, and closed-loop testing.

\section{Conclusion \& Future Work}
\label{sec:conlusion}

In this study,  we proposed and demonstrated a pipeline for editing and augmenting real-world scenes through agent-level manipulation, including cross-dataset agent injection. This capability enables the creation of rare or critical safety-related scenarios, which are essential for testing and validating AD software.
We conducted a comprehensive evaluation of state-of-the-art scene reconstruction methods, focusing on their applicability in AD and ADAS as safety-critical domains. Reconstruction fidelity for dynamic objects, especially vehicles and pedestrians, plays a vital role due to their high exposure and risk levels in traffic environments. 
Our results show that reconstruction fidelity is strongly influenced by viewpoint coverage during data collection for each object type. OmniRe consistently achieved the highest performance across all metrics and object classes, demonstrating strong robustness in novel view synthesis.Further analysis revealed a significant drop in fidelity for pedestrians at lateral distances, underscoring the need for complementary generative techniques.
We also highlighted the importance of viewpoint-aware reconstruction. Since full 360° data collection is often infeasible in real-world AD/ADAS operations, our findings emphasize the value of tagging reconstructed objects with valid visibility angles. Such metadata can inform future frameworks for automating high-fidelity scenario generation.

\section*{Acknowledgments}
This work has been partially supported by Sweden’s Innovation Agency (Vinnova, diarienummer: 2021-02585), and by the Wallenberg AI Autonomous Systems and Software Program (WASP) funded by the Knut and Alice Wallenberg Foundation.

\section*{Disclaimer}
The views and opinions expressed are those of the authors and do not necessarily reflect the official policy or position of Volvo Cars or Zenseact.


\end{document}